\documentclass{article}

\usepackage[final]{neurips_2024}

\usepackage[utf8]{inputenc} %
\usepackage[T1]{fontenc}    %
\usepackage{hyperref}       %
\usepackage{url}            %
\usepackage{booktabs}       %
\usepackage{amsfonts}       %
\usepackage{nicefrac}       %
\usepackage{microtype}      %
\usepackage{xcolor}         %

\usepackage{times}
\usepackage{latexsym}

\usepackage[T1]{fontenc}

\usepackage[utf8]{inputenc}

\usepackage{microtype}

\usepackage{inconsolata}

\usepackage{amsmath}
\usepackage[pdftex]{graphicx}
\usepackage{booktabs}

\let\cite\undefined
\newcommand{\idktoken}{\texttt{[IDK]}}
\newcommand{\idktuning}{\texttt{IDK}-tuning}
\newcommand{\idktuned}{\texttt{IDK}-tuned}
\newcommand{\idkobjective}{\texttt{IDK} objective}
\newcommand{\factor}{\textit{Uncertainty Factor}}
\newcommand{\fpreg}{$\mathcal{L}_\texttt{FP-reg}$}
\newcommand{\fpregineqn}{\mathcal{L}_\texttt{FP-reg}}

\newcommand{\mistral}{\texttt{Mistral-7B-v0.1}}
\newcommand{\bert}{\texttt{bert-base-cased}}
\newcommand{\pythias}{\mbox{\texttt{pythia-70m} -- \texttt{2.8B}}}

\usepackage{etoolbox}
\makeatletter
\patchcmd{\hyper@makecurrent}{%
    \ifx\Hy@param\Hy@chapterstring
        \let\Hy@param\Hy@chapapp
    \fi
}{%
    \iftoggle{inappendix}{%
        \@checkappendixparam{chapter}%
        \@checkappendixparam{section}%
        \@checkappendixparam{subsection}%
        \@checkappendixparam{subsubsection}%
        \@checkappendixparam{paragraph}%
        \@checkappendixparam{subparagraph}%
    }{}%
}{}{\errmessage{failed to patch}}

\newcommand*{\@checkappendixparam}[1]{%
    \def\@checkappendixparamtmp{#1}%
    \ifx\Hy@param\@checkappendixparamtmp
        \let\Hy@param\Hy@appendixstring
    \fi
}
\makeatletter

\newtoggle{inappendix}
\togglefalse{inappendix}

\apptocmd{\appendix}{\toggletrue{inappendix}}{}{\errmessage{failed to patch}}

\title{I Don't Know: Explicit Modeling of Uncertainty\\ with an \idktoken{} Token} %

\author{Roi Cohen \\
         HPI / University of Potsdam\\
        \texttt{Roi.Cohen@hpi.de} \\\And
  Konstantin Dobler \\
  HPI / University of Potsdam\\
  \texttt{Konstantin.Dobler@hpi.de} \\\And
  Eden Biran \\
  Tel Aviv University \\
  \texttt{edenbiran@mail.tau.ac.il} \\\And
  Gerard de Melo \\
  HPI / University of Potsdam\\
  \texttt{Gerard.DeMelo@hpi.de}}

\begin{document}
\maketitle

\begin{abstract}
Large Language Models are known to capture real-world knowledge, allowing them to excel in many downstream tasks. Despite recent advances, these models are still prone to what are commonly known as hallucinations, causing them to emit unwanted and factually incorrect text. In this work, we propose a novel calibration method that can be used to combat hallucinations. 
We add a special \texttt{[IDK]} (“\underline{I} \underline{d}on’t \underline{k}now”) token to the model's vocabulary and introduce an objective function that shifts probability mass to the \texttt{[IDK]} token for incorrect predictions. 
This approach allows the model to express uncertainty in its output explicitly. 
We evaluate our proposed method across multiple model architectures and factual downstream tasks.
We find that models trained with our method are able to express uncertainty in places where they would previously make mistakes while suffering only a small loss of encoded knowledge. We further perform extensive ablation studies of multiple variations of our approach and provide a detailed analysis of the precision-recall tradeoff of our method.\footnote{We release our code and \idktuned{} model checkpoints at \url{https://github.com/roi-hpi/IDK-token-tuning}.}
\end{abstract}

\section{Introduction}
\label{sec:intro}

Large Language Models (LLMs) are pretrained on massive amounts of text
to understand and generate language.
This training text includes a large portion of written human knowledge such as books, newspapers, Wikipedia, and scientific articles.
During this process, LLMs also retain a remarkable amount of the information seen during pre-training, allowing them to encode real-world knowledge in their parameters and act as knowledge bases \citep{petroni2019language, roberts2020much, cohen-etal-2023-crawling, LLMsKGs2023}.
Owing to this phenomenon, LLMs can be used in multiple settings requiring this real-world knowledge, such as closed-book question answering \citep{brown2020language, roberts2020much} and information retrieval \citep{tay2022transformer}.

\begin{figure}
\setlength{\belowcaptionskip}{-10pt}
    \centering
    \includegraphics[width=1\textwidth]{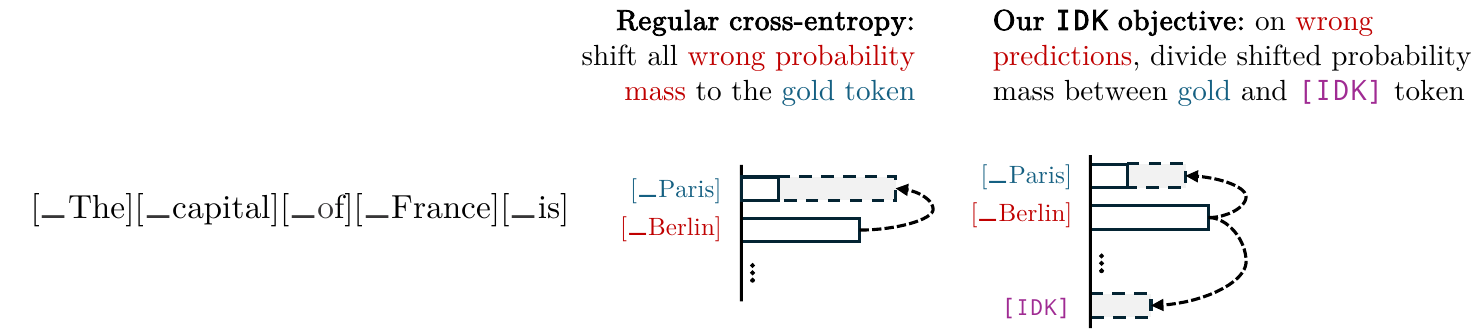}
    \caption{Illustration of our proposed \texttt{IDK} objective. During continual pretraining, we shift some probability mass of wrong predictions towards a special \idktoken{} token. The amount of shifted probability mass depends on the uncertainty in the model's prediction. We detail our method in \autoref{sec:method}.}
    \label{figure:intro}
\end{figure}

Despite the popularity of LLMs, they are prone to what is commonly referred to as hallucinations, which severely hinder their performance and reliability \citep{10.1145/3571730, ManduchiEtAl2024GenAI}. Examples of hallucinations include factually incorrect \citep{maynez-etal-2020-faithfulness, devaraj-etal-2022-evaluating, tam-etal-2023-evaluating}, inconsistent \citep{elazar-etal-2021-measuring, mundler2023self}, self-contradicting \citep{cohen2024evaluating} or non-attributable text \citep{bohnet2022attributed, rashkin2023measuring, yue-etal-2023-automatic}.

A prominent method employed to combat such hallucinations is model calibration \citep{10.5555/3305381.3305518, Brundage2020TowardTA}, which aims to calibrate the confidence of model predictions such that they are better aligned with their quality. 
This calibration allows LLMs to explicitly express uncertainty, allowing them to caveat their responses or even refrain from answering. 
Although many of the proposed methods do lead to an improvement in model calibration \citep{Geng2024ASO}, they have still been found to be lacking \citep{chen-etal-2023-close}.

In this work, we propose a novel objective function that allows LLMs to explicitly express uncertainty. We add a new special \idktoken{} (“I Don’t Know”) token to the vocabulary of the language model.
During a continued pretraining phase, we modify the conventional cross-entropy objective to express uncertainty in a next-token prediction as probability mass on the \idktoken{} token.
Specifically, each time the model fails to predict the gold label, some of the probability mass of the target is shifted to the \idktoken{} token based on an \factor{} we calculate based on the predicted logits.
We refer to our method as \idktuning{}.

Our proposed \idkobjective{} differs from previous work as we intervene during a continued pretraining phase with the language modeling task. Crucially, we do not rely on any labeled data. Moreover, this allows the model to be later finetuned on specific tasks while the model has already learned to express uncertainty.

We conduct \idktuning{} using various model architectures and sizes, and then evaluate them on diverse factual downstream tasks. Our results show a large increase in factual precision of \idktuned{} models while causing only a small decrease in recall of factual knowledge that was contained in the base model. We conduct extensive ablation studies for the individual components of our \idkobjective{} and analyze its effect on optimization dynamics. We finally show that \idktuning{} does not harm the general language modeling ability of models, such as long text generation.

In summary, our contributions include:

\begin{itemize}
    \item We propose a novel \idkobjective{} applied during pretraining which models uncertainty in a model's prediction as probability mass put on a special \idktoken{} token.
    \item We evaluate our objective using a large range of base models with different architectures and model sizes, and confirm the efficacy of \idktuning{} on a range of factual answering downstream tasks.
    \item We extensively analyze individual components of our objective and its effect on general language modeling ability.
\end{itemize}

\section{\idktuning{}}
\label{sec:method}
Our goal is to train a model to be aware of its unawareness and to effectively express it. For this, we introduce a new special token to its vocabulary: \idktoken{}. The model is intended to express uncertainty by putting probability mass on the \texttt{[IDK]} token in its predictions.
In practice, we adapt the model's pretraining objective, aiming to teach it to use the \idktoken{} token effectively. Our objective does not require annotations of uncertainty or specifically crafted datasets (e.g., Q\&A). Instead, we leverage the uncertainty captured by the pretraining objective on its pretraining data and use it to encourage probability mass on the \texttt{[IDK]} token in cases of uncertainty.
We hypothesize that this generalizes to uncertainty expressed on downstream tasks like Q\&A, which we experimentally verify later on.

We next describe in detail the technicalities of the \idktoken{} token and our training method.

\subsection{The \idktoken{} token}
The purpose of the new token is to represent lack of knowledge. 
Ideally, whenever the model would have been making a mistake, we want it to instead predict this token. 
That is, rather than generating a wrong token, we would like to model to generate the \idktoken{} token, as a means of conveying its uncertainty. We can consider this as a model expressing its lack of knowledge and may then choose to ignore its outputs. The more the model opts for this token rather than predicting the wrong answer, the more we improve the model's precision. 

For instance, let us consider the setup of Factual Sentence Completion. In this setup, the model receives an incomplete sentence as an input and is expected to complete it factually. For example, a valid input would be \texttt{``Paris is the capital of''}, and a factually correct output by the model would be \texttt{``France''}. In this setup, if the model was going to predict \texttt{``Germany''}, using the \idktoken{} token instead increases factual precision by refusing to answer a question where the answer would have been wrong. 
Naturally, almost universally predicting \idktoken{} indiscriminately may yield high precision but is not helpful. Therefore, taking into account the recall of factually correct answers is crucial in evaluating our method.  We analyze both the precision and recall of our method in \autoref{sec:results}.

We add this new \idktoken{} token to the model's vocabulary and initialize its embedding randomly. The embedding is optimized alongside the rest of the model's parameters during training.
We next describe our proposed \idkobjective{}.

\subsection{The \texttt{IDK} Training Objective}

\newcommand{\preds}{\mathbf{\hat{y}}}
\newcommand{\golds}{\mathbf{y}}

We modify the conventional cross-entropy objective between the softmax distribution over the model’s prediction and the correct answer, such that each time the model fails to predict the correct token, it is encouraged to instead put some probability mass on the \idktoken{}. 
This encouragement is modulated by an \factor{} denoted as $\lambda \in [0,1]$ that is larger the more uncertain the model is and exactly 0 when the model predicts the correct token.

We now define our modified cross-entropy objective. We use \texttt{[gold]} to denote the gold token (correct target) for each prediction. We denote the probability mass assigned to an arbitrary token \texttt{[tok]} in the prediction of a model  as $\texttt{prob}(y_t = \texttt{[tok]}| y_{<t}, x)$
We further use $\mathbf{1}_{\texttt{[IDK]}}$ to denote a one-hot target vector with one at the index of the \idktoken{} token. Per convention, $\golds$ denotes the one-hot target vector for the \texttt{[gold]} token. The modified objective is defined as follows:

\begin{equation}
    \label{eqn:loss-idk}
    \mathcal{L}_\mathrm{\texttt{IDK}} = \mathcal{L}_\mathrm{CE} (\preds,\, (1 - \lambda) \,\golds + \lambda\, \mathbf{1}_{\idktoken{}} )
\end{equation}

If the model is uncertain in its prediction, the target is shifted away from predicting the \texttt{[gold]} token and towards the \idktoken{} token. This is modulated by $\lambda$. 
Note that in case the model makes the correct prediction, $\lambda = 0$ and $\mathcal{L}_\mathrm{\texttt{IDK}}$ therefore reduces to the regular cross-entropy loss.
When the model is correct, $\mathcal{L}_\texttt{IDK}$ simply provides the signal for the correct prediction. When the model is incorrect, $\mathcal{L}_\texttt{IDK}$ provides both the signal for the correct prediction and a signal to express uncertainty.
We now detail the construction of the \factor{} $\lambda$.

\paragraph{The \factor{}.} $\lambda$ is constructed as a scalar weight with $\lambda \in [0,1]$. Intuitively, we want $\lambda$ to be close to 1 when the model is very uncertain and 0 when the model makes the correct prediction.
Based on this, we define $\lambda$ as one minus the probability mass on the gold token divided by the maximum probability mass put on any token:

\begin{equation}
\label{eqn:uncertainty_factor}
\lambda = \Pi \times \left(1 - \frac{\texttt{prob}(y_t = \texttt{[gold]} | y_{<t}, x)}{\mathrm{max_i}(\texttt{prob}(y_t = i| y_{<t}, x))}\right),
\end{equation}
where $\Pi \in [0,1]$ is a hyperparameter to control the influence of our objective.
When the gold token probability is close to the maximum probability, $\lambda$ is close to 0. If the model makes a correct prediction (the gold token is assigned the maximum probability), $\lambda$ is 0, thereby reducing \autoref{eqn:loss-idk} to the regular cross-entropy loss.
When the gold token probability is much lower than the maximum probability, $\lambda$ is close to 1, which translates to shifting almost all the probability mass of the target in \autoref{eqn:loss-idk} to the \idktoken{} token.
$\Pi$ specifies the upper bound of target probability mass that can be shifted to the \idktoken{} token. For example, $\Pi = \frac{1}{2}$ means that at most half of the probability mass in the target can be shifted to \idktoken{} while the rest remains with the gold token. In practice, we do not tune this and set $\Pi = \frac{1}{2}$. This prevents the \idktoken{} token from ever becoming a better prediction than the gold token while still providing enough signal to predict \idktoken{} for uncertain predictions. We perform an ablation of the influence of $\Pi$ in \autoref{sec:ablations}.

\paragraph{Uncertainty Regularization.}
\label{para:regularization}
An important consideration in designing the $\mathcal{L}_\texttt{IDK}$ objective is to prevent a collapse where the model is miscalibrated with too many false positive \idktoken{}s, putting too much probability mass on \idktoken{}, although it could have made the correct prediction. Therefore, we add the following anti-false positive regularization to our objective:
\begin{equation} \label{regularization_only}
\fpregineqn{} = -\mathrm{log}(1 - \texttt{prob}(y_t = \idktoken{} | y_{<t}, x)),
\end{equation}
which is exactly the binary cross-entropy objective with 0 as the target and the probability mass assigned to the \idktoken{} as the input. We only add this regularization objective when the model's prediction is correct.
This aims to minimize the \idktoken{} token's probability mass the model learns to predict in cases it knows the answer -- thus teaching it to minimize the use of this token in cases it is more certain, and is designed to reduce a decrease of its recall. We perform an ablation of \fpreg{} in \autoref{sec:ablations}.
\vspace{-2mm}
\paragraph{The final loss.} Combining all objectives, our final \idkobjective{} is therefore:

\begin{equation} \label{final}
\mathcal{L} = \begin{cases} 
\mathcal{L}_\mathrm{CE} + \fpregineqn{} & \text{if } \lambda = 0 \\
 \mathcal{L}_\mathrm{\texttt{IDK}} & \text{otherwise.} 
\end{cases}
\end{equation}

\section{Experiments}
\label{sec:experiments}

We use our proposed \idkobjective{} to tune various pretrained models to use the new \idktoken{} token.
We dub this process \idktuning{}. We then report the results of the \idktuned{} models on commonly used factual benchmarks, showing that our method improves factuality while paying only a small price in terms of knowledge recall.
We also show that model size plays a significant role in the success of our method to create an effective uncertainty-aware model.

We employ \textit{continual training} of pretrained models rather than training from scratch for two reasons: (i) the computational cost of training models that perform competitively on current benchmarks from scratch would be prohibitive, and (ii) starting from a model that is already a strong language modeler helps during the optimization process by providing a rough initial calibration that we utilize to derive the \factor{}.

\subsection{\idktuning{} Setup}
\label{sec:idktuning-setup}
We use \texttt{bert-base-cased} \citep{devlin-etal-2019-bert}, \texttt{mistralai/Mistral-7B-v0.1} \citep{jiang2023mistral}, and \mbox{\texttt{EleutherAI/pythia-70m} -- \texttt{2.8B}} \citep{10.5555/3618408.3618510} for our base models for \idktuning{}.
For \idktuning{} \texttt{Mistral-7B-v0.1}, we train on data randomly sampled from The Pile~\citep{gao2020pile}\footnote{We use \texttt{monology/pile-uncopyrighted} on the Huggingface Hub for a version of The Pile without the Books corpus, which contains copyrighted works.} with a context length of 4,096. We use example packing to fill the entire context length. We use a maximum learning rate of $4 \times 10^{-5}$ with a linear warmup for 10\% of the training steps and a cosine decay down to $2 \times 10^{-6}$. We use a batch size of 256, weight decay of 0.05, gradient clipping of 1.0 and AdamW betas (0.9, 0.95). We train for 1,024 optimizer steps resulting in a total of 1B training tokens. For the \pythias{} models, we use the same hyperparameters but reduce the context length to 2,048 to match the model's positional embeddings.
We use \texttt{bfloat16} and \texttt{float16} mixed-precision training to match \texttt{Mistral-7B-v0.1} and \texttt{pythia-410m} -- \texttt{2.8B} pretraining, respectively. For \texttt{pythia-70m}, \texttt{pythia-160m} and \texttt{bert-base-cased}, we observed \texttt{NaN} errors in the predicted logits irrespective of our loss modifications. Since the models are small enough, we switch to pure \texttt{float32} for these models without using mixed-precision. 
In addition, for \texttt{bert-base-cased} we apply MLM \citep{devlin-etal-2019-bert}, while for each input, we randomly mask one of the tokens.

\subsection{Evaluation Setup}
\label{sec:eval-setup}
\paragraph{Evaluation Data.}
We consider the following datasets: LAMA \citep{petroni2019language}, TriviaQA \citep{joshi2017triviaqa}, and PopQA \citep{mallen2022not}. These cover a wide range of queries, for example trivia questions (TriviaQA), and subject-relation-object facts phrased as queries (LAMA, PopQA). We consider the closed-book open-ended setting, where we do not provide any context or answer choices to the model. Importantly, in the case of TriviaQA and PopQA, where the input is formed as a question, we reduce it into a sentence completion task, using GPT4. Specifically, we prompt it to phrase the question as a sentence, while also providing it with some in-context examples that we manually created.
See \autoref{appx:question_rephrasing} for more details and the full prompt.
To evaluate multiple-choice question answering, we use EleutherAI's \texttt{lm-evaluation-harness}~\citep{eval-harness}. Specifically, we use ARC \citep{clark2018think}, HellaSwag \citep{zellers-etal-2019-hellaswag}, MMLU \citep{hendrycks2020measuring}, TruthfulQA \citep{lin-etal-2022-truthfulqa}, WinoGrande~\citep{10.1145/3474381}, and GSM8k \citep{Cobbe2021TrainingVT}.

\paragraph{Baselines.}
For each of the evaluation datasets, we compare the \idktuned{} model with its original base model without any further training. Furthermore, we consider three different baselines:

\begin{enumerate}
    \item \textbf{confidence-based} baseline: We use the predicted probability mass in the language modeling head of the LM as a measure of confidence in the prediction \citep{yoshikawa-okazaki-2023-selective}. We consider the first token generated by the LM. In case the corresponding probability mass of this token is greater than a fixed threshold, we consider the generation as valid. Otherwise, we consider this as an uncertainty expression (analogous to an \idktoken{} token generation in our model). To create a strong baseline, we search for the best threshold via hyperparameter tuning on the development set.

    \item \textbf{P(True)} baseline \citep{Kadavath2022LanguageM}: Given an input sentence to complete, which we refer to as $I$, we use the original model to generate the completion, which we refer to as $A$. We then concatenate $I$ and $A$ and ask the model: \emph{"Please answer either with `true' or `false' only. Is it true that: $I A$"}. If the model answer is not `true', we consider this specific example as unknown for the model -- namely the same case as if the IDK-tuned model would generate the \idktoken{}.

    \item \textbf{Semantic Entropy} baseline \citep{kuhn2023semantic, aichberger2024semantically}: We sample $K$ text generations from the model, encode them using a state-of-the-art semantic encoder and cluster their encodings. If the largest cluster size is larger than $\frac{K}{2}$, then we take a random generation out of this cluster as the model's answer. Otherwise, we consider this example as unknown. 
\end{enumerate}

\newcommand{\idkrecall}{\texttt{IDK} recall}
\newcommand{\idkfpr}{\texttt{IDK} error rate}
\paragraph{Evaluation.}
We evaluate how well our models use the new \idktoken{} token by measuring their factuality and knowledge memory, using the following metrics: (i) \textbf{Precision}: the portion of factually correct completions, out of all the claims that have been completed with any token that is different from the \idktoken{} token, i.e., the claims that the model was certain enough about, and tried to factually complete. (ii) \textbf{Recall}: the portion of factually correct completions, out of all the claims in the dataset. Namely, the portion of knowledge memory the model has, out of the entire test set we evaluate on.
(iii) \textbf{F1}: the harmonic mean of precision and recall. In the case of base models without additional calibration methods, the precision, recall, and F1-scores all correspond to their accuracy on the task. 

In \autoref{sec:ablations}, we use two further metrics to analyze the patterns when \idktuned{} models predict \idktoken{}. For this, we use the notion of correctly predicting \idktoken{}: We consider an \idktoken{} prediction to be correct if the base model does not predict the correct answer for an instance. We define (i) \textbf{\idkrecall{}}: the fraction of instances the model predicted \idktoken{} correctly out of all instances where the base model did in fact not predict the correct answer, and (ii) \textbf{\texttt{IDK} error rate}: the fraction of instances where the model predicted \idktoken{} incorrectly out of all instances where the base model did indeed predict the correct answer.

\vspace{-2mm}
\section{Results}
\label{sec:results}
\vspace{-2mm}

\begin{table*}[t]
\setlength{\belowcaptionskip}{-4pt}
\centering
\resizebox{1\linewidth}{!}{
\begin{tabular}{@{}l  ccc  ccc  ccc  ccc  ccc@{}}
\toprule
 & \multicolumn{3}{c}{LAMA Google-RE} & \multicolumn{3}{c}{LAMA T-Rex} & \multicolumn{3}{c}{LAMA SQuAD} &  \multicolumn{3}{c}{TriviaQA} & \multicolumn{3}{c}{PopQA} \\ 
\cmidrule(r){2-4}\cmidrule(lr){5-7}\cmidrule(lr){8-10}\cmidrule(lr){11-13}\cmidrule(l){14-16}
\multicolumn{1}{c}{} & \textbf{P}  & \textbf{R} & \textbf{F1}  & \textbf{P} & \textbf{R} & \textbf{F1}  & \textbf{P}  & \textbf{R} & \textbf{F1} & \textbf{P}  & \textbf{R} & \textbf{F1} & \textbf{P}  & \textbf{R} & \textbf{F1} \\
\midrule
\texttt{Mistral-7B-v0.1}      & $48.1$    &$48.1$   &$48.1$  
                    & $71.2$    &$\mathbf{71.2}$   &$71.2$ 
                    & $45.8$    &$45.8$   &$45.8$ 
                    & $52.0$    &$52.0$   &$52.0$ 
                    & $35.5$    &$\mathbf{35.5}$   &$\mathbf{35.5}$  \\
\texttt{Mistral-7B-v0.1} + The Pile  & $48.8$    &$\mathbf{48.8}$   &$48.8$  
                    & $69.9$    &$69.9$   &$69.9$ 
                    & $48.0$    &$\mathbf{48.0}$   &$48.0$ 
                    & $52.2$    &$\mathbf{52.2}$   &$52.2$ 
                    & $35.2$    &$35.2$   &$35.2$  \\
\texttt{Mistral-7B-v0.1} + Confidence Threshold    & $60.0$    &$40.0$   &$48.0$ 
                    & $80.4$    &$63.5$ &$71.0$ 
                    & $64.4$    &$33.5$ &$44.1$
                    & $70.4$    &$41.1$ &$51.9$
                    & $64.6$    &$20.6$ &$31.2$  \\ \midrule
\texttt{Mistral-7B-v0.1} + P(True)    & $54.4$    &$44.5$   &$48.9$ 
                    & $73.8$    &$65.1$ &$69.2$ 
                    & $54.9$    &$41.0$ &$46.9$
                    & $58.8$    &$47.5$ &$52.5$
                    & $40.3$    &$29.0$ &$33.7$  \\ \midrule
\texttt{Mistral-7B-v0.1} + Semantic Entropy    & $70.1$    &$38.9$   &$50.0$ 
                    & $88.0$    &$65.4$ &$75.0$ 
                    & $70.2$    &$44.5$ &$54.4$
                    & $68.5$    &$52.5$ &$59.4$
                    & $68.7$    &$20.4$ &$31.5$  \\ \midrule
\textbf{\texttt{Mistral-7B-v0.1} + \texttt{IDK}-tuning} on The Pile & $\mathbf{71.1}$    &$40.6$   &$\mathbf{51.7}$
                            & $\mathbf{88.5}$    &$65.5$   &$\mathbf{75.3}$   
                            & $\mathbf{72.0}$    &$44.3$   &$\mathbf{54.9}$
                            & $\mathbf{72.5}$    &$52.0$   &$\mathbf{60.6}$
                            & $\mathbf{78.1}$    &$20.5$   &$32.5$   \\

\bottomrule
\end{tabular}
}
\caption{Precision (P), Recall (R), and F1-scores for \texttt{Mistral-7B-v0.1}. Our \texttt{IDK}-tuning achieves the best precision with minor decreases in recall, outperforming previous work. \texttt{Mistral-7B-v0.1} + Confidence Threshold refers to the baseline based on the probability mass of the predicted answer~\citep{yoshikawa-okazaki-2023-selective}.  \texttt{Mistral-7B-v0.1} + The Pile refers to the ablation discussed in \autoref{sec:ablations}. }
\label{table:mistal_results}
\end{table*}

\begin{table*}[t]
\setlength{\belowcaptionskip}{-10pt}
\setlength\tabcolsep{2.5pt}
\footnotesize
\begin{center}
\begin{tabular}{l  ccc }
\toprule
\multicolumn{1}{c}{} & \textbf{P}  & \textbf{R} & \textbf{F1} \\
\midrule
\texttt{Mistral-7B-v0.1}      & $28.2$    &$\mathbf{28.2}$   &$28.2$  \\
\texttt{Mistral-7B-v0.1} + The Pile      & $28.3$    &$28.3$   &$28.3$  \\
\texttt{Mistral-7B-v0.1} + Confidence Threshold    & $45.0$    &$18.5$   &$26.2$ \\ \midrule
\textbf{\texttt{Mistral-7B-v0.1} + \texttt{IDK}-tuning} on The Pile   & $\mathbf{48.8}$    &$20.8$   &$ \mathbf{29.2}$  \\

\bottomrule
\end{tabular}
\end{center}
\caption{Precision (P), Recall (R), and F1-scores of our model on the \texttt{lm-eval-harness}, compared to baselines.}
\label{table:harness_results}
\end{table*}

We next report results showing that our proposed \idktuning{} method can effectively improve factuality while causing only a small loss of existing knowledge.

\subsection{Main Results}
\label{sec:main-results}
\paragraph{\mistral{} results.} \autoref{table:mistal_results} shows the results of our largest model \mistral{} on factual closed-book sentence completion datasets. Our results show that the \idktuned{} \mistral{} has a much higher precision -- namely the model generates significantly fewer factually incorrect completions and instead puts probability mass on the \idktoken{} token. However, the model does show decreased knowledge recall on some tasks. Overall, we observe an increase in the average F1-score. 
\autoref{table:harness_results} shows the averaged results on the \texttt{lm-eval-harness} datasets.
The trend here is similar, although the increase in precision compared to baselines is slightly lower. 
This suggests that the model tends to be more certain when it comes to multiple-choice questions.

\begin{figure}
    \centering
    \includegraphics[width=0.8\textwidth]{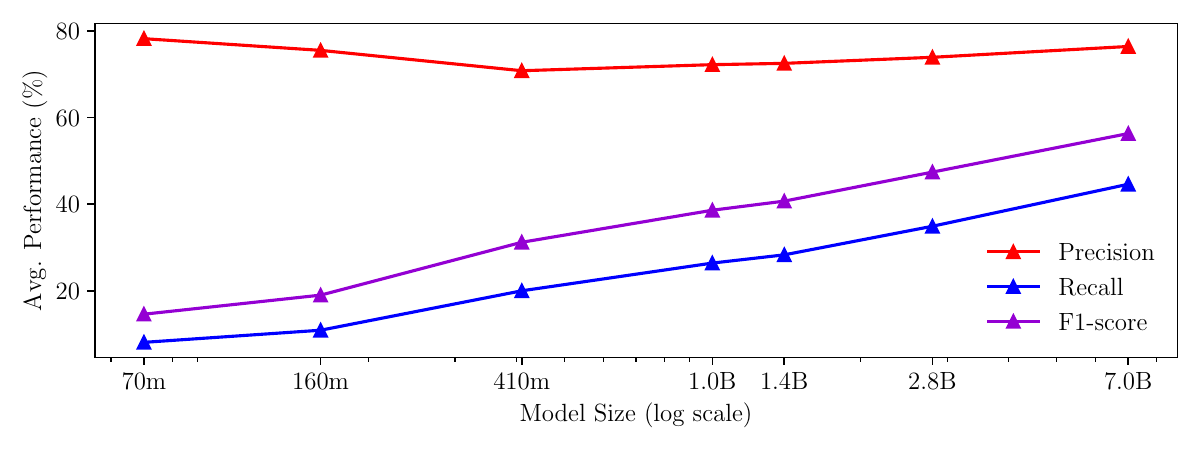}
    \vspace{-5mm}
    \caption{Average performance on closed-book factual sentence completion benchmarks of \idktuned{} models in terms of their parameter count. 70m to 2.8B are \pythias{}, while 7.0B is \mistral{}.}
    \label{fig:pythia-scaling}
\end{figure}

\paragraph{Scaling behavior of \idktuning{}.}
We further investigate the effect of model size on the success of \idktuning{}. We conduct \idktuning{} for each of the \pythias{} models as detailed in \autoref{sec:idktuning-setup}. In \autoref{fig:pythia-scaling}, we plot the average precision, recall, and F1-score for each of \pythias{} as well as \mistral{}, over all the closed-book sentence completion datasets. We observe a clear trend of recall and F1-score increasing log-linearly with the model size. The precision of \idktuned{} models increases only slightly as the model size increases. 
For the two smallest models we investigate (\texttt{pythia-70m} and \texttt{pythia-160m}), our method is arguably not effective, as the \idktuned{} model's recall collapses (we further analyze this in \autoref{sec:training_collapse}).

\paragraph{\texttt{bert-base-cased} results.}

\begin{table}[t]
\begin{center}
\resizebox{1\linewidth}{!}{

\begin{tabular}{l  ccc  ccc  ccc }
\toprule
 & \multicolumn{3}{c}{LAMA Google-RE} & \multicolumn{3}{c}{LAMA T-Rex} & \multicolumn{3}{c}{LAMA SQuAD} \\
 \cmidrule(r){2-4}\cmidrule(lr){5-7}\cmidrule(lr){8-10}
\multicolumn{1}{c}{} & \textbf{P}  & \textbf{R} & \textbf{F1}  & \textbf{P} & \textbf{R} & \textbf{F1}  & \textbf{P}  & \textbf{R} & \textbf{F1} \\
\midrule
\texttt{bert-base-cased}      & $23.0$    &$\mathbf{23.0}$   &$23.0$   
                    & $59.8$    &$\mathbf{59.8}$   &$59.8$ 
                    & $9.5$    &$\mathbf{9.5}$   &$9.5$  \\
\texttt{bert-base-cased}  + Confidence Treshold    & $58.8$    &$15.8$   &$24.9$ 
                    & $71.5$    &$35.9$ &$47.8$
                    & $69.5$    &$5.0$ &$9.3$  \\ \midrule
\textbf{\texttt{bert-base-cased} + \idktuning{}}   & $\mathbf{78.1}$    &$15.9$   &$\mathbf{26.4}$
                            & $\mathbf{72.5}$    &$53.0$   &$\mathbf{61.2}$
                            & $\mathbf{80.2}$    &$6.4$   &$\mathbf{11.9}$  \\

\bottomrule
\end{tabular}
}
\end{center}
\caption{Precision (P), Recall (R), and F1 scores for of our \texttt{IDK}-tuned \bert{} on the evaluation benchmarks, compared to baselines.}
\label{table:bert_results}
\end{table}

\autoref{table:bert_results} reports the results of out \idktuned{} \bert{} model. We see a similar trend as in our evaluation of \mistral{}. Factuality is improved, while recall is reduced by only a small amount.

\subsection{Ablations}
\label{sec:ablations}
We perform an ablation study of our method to further investigate the effectiveness of each of its components. For our study, we calculate the \idkrecall{} and \idkfpr{} on the closed-book factual sentence completion datasets. We study the effect of $\Pi$, $\lambda$ and the \fpreg{} term. For this, we perform \idktuning{} using \mistral{} with the same hyperparameters as our main runs with different combinations of the studied components\footnote{Due to computational constraints, we run this for a reduced set of $\Pi$ for the cases with adaptive $\lambda$.}.
We plot the \idkrecall{} for different values of $\Pi$ in \autoref{figure:pi_ablation}. In \autoref{figure:regularization_factor_ablation}, we plot the \idkrecall{} vs.\ \idkfpr{} tradeoff. \idkrecall{} and \idkfpr{} are defined in \autoref{sec:eval-setup}. We study different aspects of these results below:

\begin{figure}
    \centering
    \begin{minipage}{0.49\textwidth}
      \centering
    \includegraphics[width=1\textwidth]{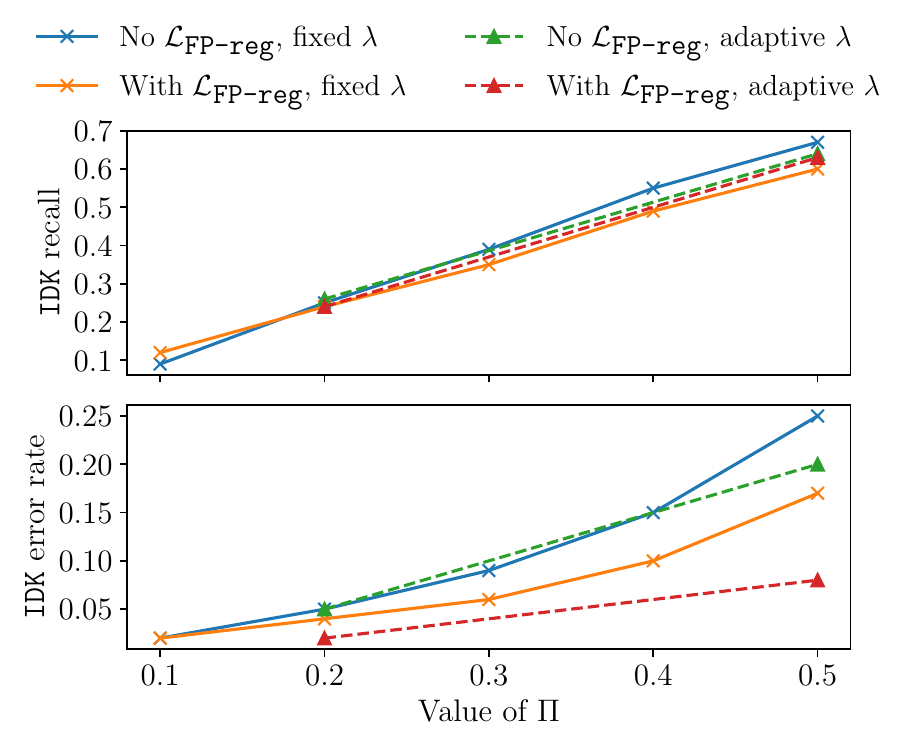}
    \caption{Ablation study of different values for the $\Pi$ factor that controls the upper bound of probability mass put on \idktoken{} in the target.}
    \label{figure:pi_ablation}
\end{minipage}
\hfill
\begin{minipage}{0.49\textwidth}
    \centering
    \includegraphics[width=1\textwidth]{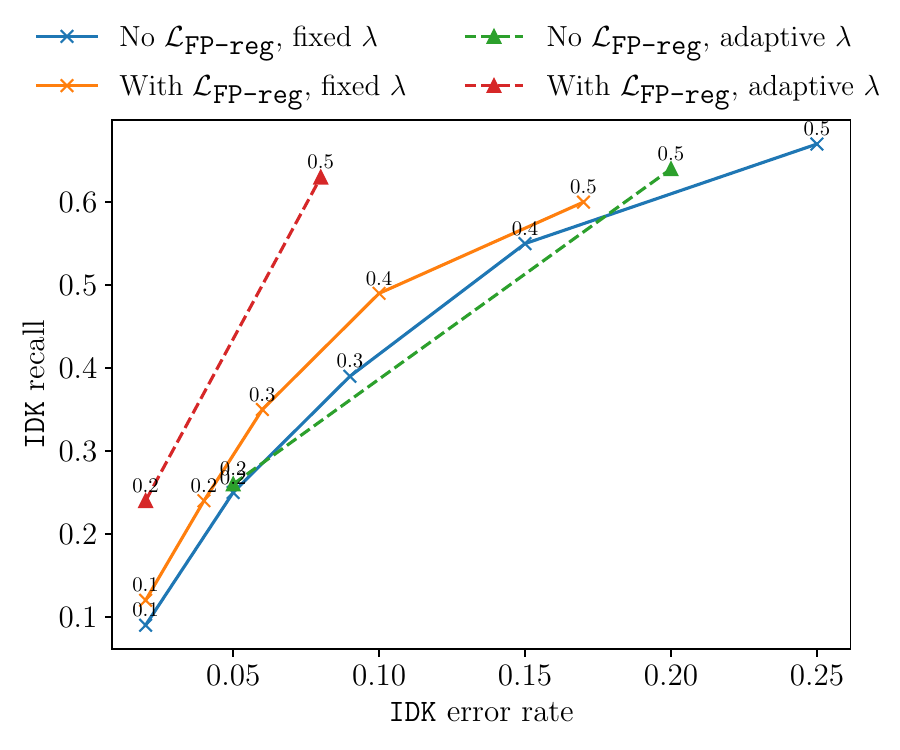}
    \caption{Tradeoff between \idkrecall{} and \idkfpr{} for different parameter combinations. We annotate each data point with its corresponding $\Pi$ value.}
    \label{figure:regularization_factor_ablation}
\end{minipage}
\end{figure}

\paragraph{1. Analysis of the adaptive nature of the \factor{} $\lambda$.} The \factor{} $\lambda$ defined in \autoref{eqn:uncertainty_factor} is \textit{adaptive}, meaning the amount of probability mass shifted to \idktoken{} depends on the predicted probability distribution. Another possible choice is to use a \textit{fixed} $\lambda \in [0, 1]$. We analyze this in \autoref{figure:pi_ablation} and \autoref{figure:regularization_factor_ablation}\footnote{For a fixed $\lambda$, we set $\lambda = \Pi$.}. We can see that using the adaptive $\lambda$ formulation results in a lower \idkfpr{} without a major decrease in \idkrecall{}.
\vspace{-1mm}
\paragraph{2. Effect of the \fpreg{} regularization.}
We also study the effect of the \fpreg{} term (see ~\autoref{para:regularization}). 
Again, we see that using \fpreg{} results in a reduced \idkfpr{} without decreasing \idkrecall{} significantly.
\vspace{-1mm}
\paragraph{3. Effect of the upper bound hyperparameter $\Pi$.} 
We also study the effect of $\Pi$, which is the upper bound of the \factor{} (see \autoref{eqn:uncertainty_factor}). 
Our ablation study demonstrates that increasing $\Pi$ results in an increase in correct predictions of \idktoken{} (higher \idkrecall{}), at the cost of a small increase of erroneous \idktoken{} predictions (\idkfpr{}). The \idkfpr{} increases less when using both our proposed adaptive $\lambda$ and \fpreg{}.

\paragraph{Effect of knowledge contained in The Pile.} Since we conduct further pretraining on The Pile, improved performance of our method could be partly explained by additional knowledge that the model learns during \idktuning{}. However, we show that this is not the case. In the case of the \pythias{} models, our data used for \idktuning{} exactly matches their pretraining data. For \texttt{Mistral-7B-v0.1}, this is not known although The Pile was likely also included. We note that the language modeling performance on The Pile of our models during \idktuning{} actually very slightly decreases rather than improving, suggesting the absence of any newly learned knowledge. However, to completely rule out any such effects, we trained \texttt{Mistral-7B-v0.1} on the exact sample of The Pile used for \idktuning{} but with the regular cross-entropy objective. We report the performance of this model in \autoref{table:mistal_results}. Indeed, \texttt{Mistral-7B-v0.1} with further training on The Pile performs similarly to the base \texttt{Mistral-7B-v0.1} on average.

\subsection{Analysis of Optimization Stability}
\vspace{-1mm}
\paragraph{Collapse to \idktoken{}.}Highly optimizing every component of the standard language modeling task with Transformers has made it easy to forget that optimization processes of deep neural networks can be brittle and divergent. Naively replacing the regular cross-entropy objective with our \texttt{IDK}-loss $\mathcal{L}_\texttt{IDK}$ leads to a collapse of training where the model simply always learns to put most probability mass on the \idktoken{}. We already account for this by (i) introducing the $\Pi$ inhibitor to allow us to set an upper bound on the maximum probability mass that is assigned to the \idktoken{} in the target vector and (ii) introducing the additional \fpreg{} regularization to provide an additional signal that punishes probability mass being assigned to \idktoken{} when the model's prediction is already correct. 

In practice, we see that the regular cross-entropy loss shows a small uptick at the very beginning of \idktuning{}. In almost all runs, this recovers quickly back to baseline levels, where it remains. We find that with $\Pi = 0.5$ and the \fpreg{} regularization, most training runs are stable without further model-specific tuning. 
\vspace{-1mm}
\paragraph{Collapse for small models \texttt{pythia-70m} and \texttt{pythia-160m}.}
\label{sec:training_collapse}
However, for \texttt{pythia-160m} and \texttt{pythia-70m}, which are the only runs in our experiments that diverge even with our added regularization losses, the regular cross-entropy keeps on rising with a large spike. Concretely, the predicted distributions not only show an increased cross-entropy with the targets but also a sharply increasing entropy: we observe that the predicted distributions collapse towards a uniform distribution. At the worst point, 0\% of the predictions of \texttt{pythia-160m} are correct. However, both models somewhat recover towards the end of training but stay well below baseline levels in terms of language modeling performance. We note that this is a different collapse pattern than the collapse towards almost always predicting \idktoken{} observed without our regularization terms. 

We further analyzed this and observe that for both \texttt{pythia-160m} and \texttt{pythia-70m}, the initial probability mass assigned to the \idktoken{} token is so small that it gets rounded to zero even when using \texttt{float32} precision. This causes the $\mathcal{L}_\texttt{IDK}$ loss to be very large, resulting in large gradient norms. Already for \texttt{pythia-410m}, the initial probability mass on \idktoken{} is substantial enough to prevent this (albeit still a very small value smaller than 5 $\times 10^-9$). Both \texttt{pythia-160m} and \texttt{pythia-70m} also show a larger initial entropy in their predicted distributions (i.e., ``flatter'' predicted distributions). We conjecture that an adapted initialization of the \idktoken{} token and/or a small bias towards \idktoken{} at the beginning of training could prevent this divergence. As we only encounter this issue for the small \texttt{pythia-160m} and \texttt{pythia-70m} models, we leave further investigation of this for future work.
\vspace{-1mm}
\subsection{Text Generation}
\label{sec:summarization}
\vspace{-1mm}

\begin{table*}[t]
\setlength{\belowcaptionskip}{-4pt}
\centering
\resizebox{1\linewidth}{!}{
\begin{tabular}{@{}l  c  c  c@{}}
\toprule
 & \multicolumn{1}{c}{Legal Plain English} & \multicolumn{1}{c}{TLDR} & \multicolumn{1}{c}{SPEC5G} \\ 
\midrule
\texttt{Mistral-7B-v0.1}      & $\mathbf{17.5}$    
                    & $\mathbf{14.1}$     
                    & $37.2$       \\
\texttt{Mistral-7B-v0.1} + The Pile  & $17.4$  
                    & $\mathbf{14.1}$   
                    & $\mathbf{37.3}$      \\
\textbf{\texttt{Mistral-7B-v0.1} + \texttt{IDK}-tuning} on The Pile & $17.2$  
                            & $14.0$     
                            & $36.9$    \\

\bottomrule
\end{tabular}
}
\caption{RougeL scores on different summarization tasks to measure the impact of \idktuning{} on other language model abilities. \texttt{Mistral-7B-v0.1} + The Pile refers to the ablation discussed in \autoref{sec:ablations}. }
\label{table:summarization_results}
\end{table*}

To assess whether our \idktuning{} might harm other different downstream language skills, which are not necessarily only factual, we evaluate the \idktuned{} \mistral{} on the task of text summarization, and compare its results to those of the original model. For this, due to the high likelihood of the \idktoken{} token being generated during a longer text generation process, we use greedy decoding and ignore the \idktoken{} token. For this experiment, we use four different common summarization benchmarks: Legal Plain English \citep{manor-li-2019-plain}, TLDR \citep{volske-etal-2017-tl}, and SPEC5G \citep{karim-etal-2023-spec5g}. We measure performance using RougleL \citep{lin-2004-rouge}, as it is widely used in related work, and report the results in \autoref{table:summarization_results}. The \idktuned{} \mistral{} performs only slightly worse than the original base model. This is an encouraging result, as it means that \idktuning{} does not necessarily harm other language skills of pretrained language models.  

\subsection{Error Analysis}
\label{sec:error_analysis}
\vspace{-1mm}

\begin{table}[t]
    \centering
    \footnotesize
    \begin{tabular}{lcccc}
    \toprule
    & No effect & Noise & White Noise & Abstaining  \\ \midrule
         \mistral{}
         &  68.5\% & 9\% & 6.5\% & 16\% \\ \midrule
          \texttt{pythia-2.8B}
         &  59.5\% & 13.5\% & 12.5\% & 14.5\%  \\ \midrule
          \texttt{pythia-70m}
         &  52\% & 18.5\% & 22\% & 7.5\% \\
         \bottomrule
    \end{tabular}
    \vspace{3mm}
    \caption{Error type distribution on 200 failures of our IDK-tuned models.}
\label{table:error_analysis}
\end{table}
To gauge the effect of \idktuning{} on model responses to factual prompts and questions, we conduct an in-depth manual analysis on a random sample of 200 (40 from each dataset) of the model's incorrect generations (generations that do not contain the correct answer). We conduct this analysis for three models across model sizes: \texttt{pythia-70m}, \texttt{pythia-2.8B}, and \mistral{}. We then categorize each of these incorrect generations to one of the following categories:
\begin{enumerate}
    \item \emph{No Effect}: Both the original model and the \idktuned{} model generate the same (incorrect) answer.
    \item \emph{Noise}: The original model generates the correct answer, while the \idktuned{} model does not.
    \item \emph{White Noise}: Both the original and \idktuned{} models do not generate the correct answer, however the \idktuned{} model generates a different one.
    \item \emph{Abstain}:  The \idktuned{} model abstains from answering by generating text such as ``unknown" or ``mystery".
\end{enumerate}

The results are shown in \autoref{table:error_analysis}. Our analysis suggest that first, the bigger the model, the fewer changes our training approach causes in the model’s generations, and second, the bigger the model, the greater its ability to abstain from answering via words (which generally can be interpreted as equal to generating an \idktoken{} token, although harder to evaluate automatically).

\vspace{-1mm}
\section{Related Work}

\paragraph{Model Calibration.}
\label{sec:related_calib}
Our goal is closely related to the key challenge of model calibration~\citep{pmlr-v70-guo17a}: to provide a measure of the probability that a prediction is incorrect alongside the actual prediction. The problem of factual error detection can be viewed as a variation of calibration, where instead of a continuous probability, we provide a binary prediction for whether the model is correct or not. This is also related to the setting of selective prediction, where models can abstain from answering a query \citep{varshney-etal-2022-investigating, kamath2020selective}.
Common approaches to calibration are to perform various transformations on a model's output logits \citep{desai2020calibration, jiang-etal-2021-know}, and measuring uncertainty \citep[e.g., see][]{kuhn2023semantic}. 
More recent works have studied the use of LMs for providing calibration, by training them on statements known to be factually correct or incorrect. This ``supervised'' approach has been explored via fine-tuning \citep{Kadavath2022LanguageM, lin2022teaching}, in-context learning \citep{cohen-etal-2023-crawling, alivanistos2022prompting}, zero-shot instruction-oriented \citep{cohen-etal-2023-lm} and consistency sampling \citep{yoran-etal-2023-answering} techniques.
A more recent work \citep{azaria-mitchell-2023-internal} uses the internal state of the model for classifying whether it is certain or not. Our work builds upon this, aiming to teach the model to assess and express its own uncertainty via the new \idktoken{} token we introduced.

\paragraph{Attribution.} 
 Another related line of work focuses on checking whether LM-generated texts are faithful to a given source text \citep{bohnet2022attributed, honovich2022true}. This problem has been addressed via several approaches, including question generation \citep{wang-etal-2020-asking, honovich-etal-2021-q2, scialom-etal-2021-questeval}, NLI \citep{thorne-etal-2018-fact, welleck-etal-2019-dialogue, maynez-etal-2020-faithfulness, dziri-etal-2022-evaluating, gao2022rarr, kamoi-etal-2023-shortcomings}, data augmentation \citep{atanasova-etal-2022-fact, wright-etal-2022-generating, gekhman2023trueteacher}, and planning schemes that allow the model to self-edit its own generation \citep{schick2022peer}.
Unlike these works, we are not assuming any reference text or external knowledge bases.
Instead, we aim to teach the model to decide on its own whether it is likely to be able to factually complete a sentence correctly.

\vspace{-1mm}
\section{Conclusion}
\label{sec:conclusion}

We propose a novel method for improving LMs' factuality by adding a special \idktoken{} token to an LM's vocabulary.
Alongside the new \idktoken{} token, we introduce a novel pretraining objective called \idktuning{} to model uncertainty in the model's prediction as the probability mass assigned to the \idktoken{}.
Crucially, \idktuning{} requires no labeled data and is instead a drop-in replacement of the conventional cross-entropy loss used for self-supervised language modeling on web-crawled texts.
This allows us to explore uncertainty-aware training at a large scale. In our experiments, we conduct {continued pretraining} of a diverse range of pretrained models using the \idkobjective{}. 

Evaluation on factual sentence completion and multiple-choice benchmarks shows that \idktuned{} models can complete these tasks with much higher precision by refusing to answer (assigning high probability mass to the \idktoken{} token) in cases when the base model would have given a wrong answer.
This comes at only small decreases in recall.
We investigate the scaling behavior of our method with respect to model size using the Pythia model suite~\citep{10.5555/3618408.3618510}, perform several ablation studies for individual components of our \idkobjective{}, and verify that the general language modeling ability of \idktuned{} models does not degrade.

Our work can be extended in several ways. For example, since we do not rely on any labels of our training data used for \idktuning{}, we potentially apply our objective for next-token predictions where it might be ill-posed. Instead, we can perform lightweight filtering of relevant next-token predictions, such as named entities, focusing our objective more on factual next-token predictions. Also, \idktuning{} can be applied during pretraining from scratch, where our \texttt{IDK} objective could have interesting interactions with the acquisition of new knowledge during this stage.

\vspace{-1mm}
\section{Limitations}
\label{sec:limitations}

We note a few limitations of our proposed method. First, it requires a full pretraining of LMs on relatively large corpus. This of course is both highly computationally expensive and time-consuming. It is likely often the case that this kind of training cannot be conducted on typical  academic lab resources, on a large enough model, in a reasonable amount of time.

Second, as discussed in \autoref{sec:summarization}, our method may slightly harm certain language skills, such as long text generation. Other downstream skills may be affected more significantly. We further discuss potential risk and biases in \autoref{sec:impact}.

\vspace{-1mm}
\section*{Acknowledgements}
Roi Cohen and Gerard de Melo received funding from The Goldman Sachs Group, Inc., New York, NY, USA. Konstantin Dobler thanks the German Federal Ministry for Education and Research (BMBF) through the project «KI-Servicezentrum Berlin Brandenburg» (01IS22092) and the European Laboratory for Learning and Intelligent Systems (ELLIS) PhD program for support.
We further express our gratitude to the NeurIPS 2024 reviewers for their helpful comments.

\bibliographystyle{plainnat}
\bibliography{anthology,custom}

\appendix

\section{Impact}
\label{sec:impact}
As discussed in \autoref{sec:intro}, one of the main disadvantages of current LMs is their tendency to factually mislead the user by generating factual incorrect statements. Hence, the main impact of our work is to reduce such factual mistakes via our proposed method. Still, it is evident that this sort of approach can by no means completely eliminate hallucinations.
It is important to stress that we propose a single method, not a system design for safe deployment of LLMs. In practice, we anticipate our method to be coupled with other checks and balances, forming a safe system.

Additionally, in this work, we use The Pile as a dataset to train models. The Pile is a web-crawled corpus, which likely harbors text reflecting various forms of biases. One impact of applying \idktuning{} is that the model may learn to answer in a biased way if this bias appears in its training data, while avoiding answers that rarely appear in its training data. This shows the need for more research on compiling high-quality training corpora.

\section{Computational Resources}
\label{app:comp-resources}
For \idktuning{} of \mistral{}, we use Nvidia H100 or A100 GPUs depending on availability. For \idktuning{} \pythias{}, we use 1-4 Nvidia A6000 GPUs. 
For \idktuning{} of \bert{}, we use a single Nvidia A100 GPU.

\section{Questions Rephrasing}
\label{appx:question_rephrasing}
As mentioned in \autoref{sec:eval-setup}, for TriviaQA and PopQA, where the input is formed as a question, we reduce each of these input examples into a sentence completion task input, using GPT4. If we denote a random input question from one of these datasets by $x$, then our prompt to GPT4 is the following:

{\fontfamily{qcr}\selectfont
Please rephrase the following question as an input for a sentence completion task. 
For example: 

For the question: "Where was Michael Jackson born?", the sentence should be: "Michael Jackson was born in". 

For the question: "Who is Barack Obama's wife", the sentence should be: "The wife of Barack Obama is". 

For the question: "Where in England was Dame Judi Dench born?", the sentence should be:
}

We found this prompt to be effective enough after manually testing it on a development set of a 45 examples.

\newpage
\section*{NeurIPS Paper Checklist}

\begin{enumerate}

\item {\bf Claims}
    \item[] Question: Do the main claims made in the abstract and introduction accurately reflect the paper's contributions and scope?
    \item[] Answer: \answerYes{} %
    \item[] Justification: For all claims made in the abstract and introduction, we provide experimental results that back these claims.

\item {\bf Limitations}
    \item[] Question: Does the paper discuss the limitations of the work performed by the authors?
    \item[] Answer: \answerYes{} %
    \item[] Justification: See \autoref{sec:limitations}.

\item {\bf Theory Assumptions and Proofs}
    \item[] Question: For each theoretical result, does the paper provide the full set of assumptions and a complete (and correct) proof?
    \item[] Answer: \answerNA{} %
    \item[] Justification: We do not include new theoretical results that warrant proofs.

    \item {\bf Experimental Result Reproducibility}
    \item[] Question: Does the paper fully disclose all the information needed to reproduce the main experimental results of the paper to the extent that it affects the main claims and/or conclusions of the paper (regardless of whether the code and data are provided or not)?
    \item[] Answer: \answerYes{} %
    \item[] Justification: We provide the formulation of our objective in \autoref{sec:method} and hyperparameters as well as further details to reproduce our trainings in \autoref{sec:idktuning-setup}.

\item {\bf Open access to data and code}
    \item[] Question: Does the paper provide open access to the data and code, with sufficient instructions to faithfully reproduce the main experimental results, as described in supplemental material?
    \item[] Answer: \answerNo{} %
    \item[] Justification: We will publish all datasets, code, and model checkpoints with camera-ready version.

\item {\bf Experimental Setting/Details}
    \item[] Question: Does the paper specify all the training and test details (e.g., data splits, hyperparameters, how they were chosen, type of optimizer, etc.) necessary to understand the results?
    \item[] Answer: \answerYes{} %
    \item[] Justification: We provide these details in \autoref{sec:idktuning-setup}. See also our answer to question 4.

\item {\bf Experiment Statistical Significance}
    \item[] Question: Does the paper report error bars suitably and correctly defined or other appropriate information about the statistical significance of the experiments?
    \item[] Answer: \answerNo{} %
    \item[] Justification:  
    Our evaluations are done via prompting rather than fine-tuning (see \autoref{sec:eval-setup}), yielding no source of randomness to aggregate into error bars. Our large-scale continual training experiments are, unfortunately, too expensive to repeat multiple times with different random seeds.

\item {\bf Experiments Compute Resources}
    \item[] Question: For each experiment, does the paper provide sufficient information on the computer resources (type of compute workers, memory, time of execution) needed to reproduce the experiments?
    \item[] Answer: \answerYes{} %
    \item[] Justification: We provide those details in APPENDIX (see \autoref{app:comp-resources}).
    
\item {\bf Code Of Ethics}
    \item[] Question: Does the research conducted in the paper conform, in every respect, with the NeurIPS Code of Ethics \url{https://neurips.cc/public/EthicsGuidelines}?
    \item[] Answer: \answerYes{} %
    \item[] Justification: We carefully read the NeurIPS Code of Ethics document and made sure it's aligned with our work. One potential impact is discussed in \autoref{sec:impact}.

\item {\bf Broader Impacts}
    \item[] Question: Does the paper discuss both potential positive societal impacts and negative societal impacts of the work performed?
    \item[] Answer: \answerYes{}{} %
    \item[] Justification: We discuss the potential social impacts in \autoref{sec:impact}.
    
\item {\bf Safeguards}
    \item[] Question: Does the paper describe safeguards that have been put in place for responsible release of data or models that have a high risk for misuse (e.g., pretrained language models, image generators, or scraped datasets)?
    \item[] Answer: \answerNA{} %
    \item[] Justification: We only continually pretrain already public models on up to 1B tokens. We believe that the resulting checkpoints do not warrant additional safeguards.

\item {\bf Licenses for existing assets}
    \item[] Question: Are the creators or original owners of assets (e.g., code, data, models), used in the paper, properly credited and are the license and terms of use explicitly mentioned and properly respected?
    \item[] Answer: \answerYes{} %
    \item[] Justification: In \autoref{sec:experiments}, we mention and cite each of the models, datasets, and training technuiqs we used in our work.

\item {\bf New Assets}
    \item[] Question: Are new assets introduced in the paper well documented and is the documentation provided alongside the assets?
    \item[] Answer: \answerYes{} %
    \item[] Justification: Yes, we properly explained each of the new assets we introduced. Additionally, in \autoref{appx:question_rephrasing}, we provide the complete prompt we used in order to create our closed-booked sentence completion dataset as discussed in \autoref{sec:eval-setup}.

\item {\bf Crowdsourcing and Research with Human Subjects}
    \item[] Question: For crowdsourcing experiments and research with human subjects, does the paper include the full text of instructions given to participants and screenshots, if applicable, as well as details about compensation (if any)? 
    \item[] Answer: \answerNA{} %
    \item[] Justification: No crowdsourcing experiments with human subjects were conducted.

\item {\bf Institutional Review Board (IRB) Approvals or Equivalent for Research with Human Subjects}
    \item[] Question: Does the paper describe potential risks incurred by study participants, whether such risks were disclosed to the subjects, and whether Institutional Review Board (IRB) approvals (or an equivalent approval/review based on the requirements of your country or institution) were obtained?
    \item[] Answer: \answerNA{} %
    \item[] Justification: No study with human participants was conducted.

\end{enumerate}
\end{document}